\documentclass[runningheads]{llncs}
\usepackage{graphicx}
\usepackage{cite}
\usepackage{float}
\usepackage{subcaption}

\usepackage{amsmath,amsfonts,amsthm,bm}
\usepackage{adjustbox}
\usepackage{array,multirow}

\begin{document}
\title{Domain Adaptive Egocentric Person Re-identification}
\titlerunning{Domain Adaptive Egocentric Re-ID}
%
\author{Ankit Choudhary\inst{1}\and
Deepak Mishra \inst{1}\and
Arnab Karmakar \inst{1,2}
}
\authorrunning{A. Choudhary et al.}
%
\institute{Indian Institute of Space Science and Technology, Thiruvananthapuram, India - 695547\\
\email{choudhary.ankit00@gmail.com}\\\email{deepak.mishra@iist.ac.in}
\and
Human Space Flight Center, ISRO, Bengaluru, India - 560054\\
\email{arnabk-hsfc@isro.gov.in}
}
\maketitle              
\begin{abstract}
Person re-identification (re-ID) in first-person (egocentric) vision is a fairly new and unexplored problem. With the increase of wearable video recording devices, egocentric data becomes readily available, and person re-identification has the potential to benefit greatly from this. However, there is a significant lack of large scale structured egocentric datasets for person re-identification, due to the poor video quality and lack of individuals in most of the recorded content. Although a lot of research has been done in person re-identification based on fixed surveillance cameras, these do not directly benefit egocentric re-ID. Machine learning models trained on the publicly available large scale re-ID datasets cannot be applied to egocentric re-ID due to the dataset bias problem. The proposed algorithm makes use of neural style transfer (NST) that incorporates a variant of Convolutional Neural Network (CNN) to utilize the benefits of both fixed camera vision and first-person vision. NST generates images having features from both egocentric datasets and fixed camera datasets, that are fed through a VGG-16 network trained on a fixed-camera dataset for feature extraction. These extracted features are then used to re-identify individuals. The fixed camera dataset Market-1501 \cite{market} and the first-person dataset EGO Re-ID \cite{21} are applied for this work and the results are on par with the present re-identification models in the egocentric domain.

\end{abstract}
\section{Introduction}
With the introduction of wearable cameras in the 1990s by Steve Mann, a new field of egocentric vision has opened up in the computer vision community. Due to the affordability of these first-person recording devices, huge data is being generated almost every day. However, the currently available structured datasets for egocentric vision are quite small, due to the inherent challenges of first-person vision such as blurriness due to motion of the camera, illumination change, and poor video quality. Also, most of these data cannot be used for person re-identification tasks due to a lack of individuals in these videos, creating a lack of dedicated egocentric videos that can be used for the task of egocentric re-ID.\par
Ample research has been carried out in the field of fixed-camera based re-ID, but their applicability is limited for egocentric re-ID. A re-ID model trained on a third person re-ID dataset tends to learn the intricate features of that specific dataset only, but it becomes obsolete when re-identifying individuals in an egocentric dataset. The existing literature for first-person re-ID i.e. body parts based \cite{21} and multiple views based\cite{19} methods also suffer from dataset bias. \par
To overcome these challenges, we propose a neural style transfer (NST) based domain adaptation technique for egocentric person re-identification, which make use of present fixed camera re-ID datasets to improve the performance of egocentric re-ID. The application of domain adaptation to fixed camera datasets has been widely studied in literature \cite{23}\cite{24}. But the idea of using domain adaptation to bridge the gap between fixed camera datasets and first-person datasets have never been studied before to the best of our knowledge.\par
In this work, the following contributions are made to the research on first-person re-identification: (1)	we use style-transferred images generated by NST to develop a re-ID model which can accurately re-identify individuals from first-person vision dataset even when trained on a fixed-camera dataset (i.e., domain adaptation), thus decreasing/eliminating dataset bias; (2) we compare and contrast the accuracies of our re-ID model on the generated (or style-transferred) images as well as normal first-person images.\par

\section{Related Work}
\subsection{Classic Person Re-ID}
The initial developments in person re-ID make use of handcrafted features such as texture and color. Gray et al. \cite{3} used color channels along with texture filters to re-identify pedestrian images partitioned into six horizontal stripes. Color histograms and moments are extracted followed by dimension reduction using PCA and LFDA in Pedagadi et al. \cite{4}. With the onset of deep learning, CNNs have found wide applications in person re-ID. In \cite{10}, long short-term memory (LSTM) modules are utilized through a Siamese network. Gating functions after each convolutional layer are used in \cite{11} to capture discriminative patterns in re-ID.

\subsection{Egocentric Person Re-ID}
Person re-ID in first-person videos have gained popularity with the widespread use of first-person recording devices. Visual and sensor metadata were used in \cite{18} to successfully perform re-ID. Local features are extracted followed by the application of 3D convolution to encode temporal information. In \cite{19}, facial features were extracted and similarities were computed between a pair of cameras to yield optimal and consistent re-ID results. People are identified by dividing their images into meaningful body parts in \cite{21}. Furthermore, the contribution of different body parts is calculated by taking into account human gaze information.\par
Our work utilizes an egocentric dataset shot through a single camera in \cite{21}. Rather than depending on different body parts, we take into account the complete image of a person while re-identifying him.

\subsection{Domain Adaptation in Image-to-Image Translation and Person Re-ID}
Domain adaptation has led to significant success in image-to-image translation through GANs \cite{45} and CNNs \cite{39}. pix2pix framework for image-to-image translation was proposed in \cite{40}. A conditional GAN was used to learn mappings between two different sets of images, i.e., source images and target images. Adapting simulation data to real-world data was done through domain confusion loss and pairwise loss in \cite{41}. Gatys et al. \cite{50} computed the differences between the style and content of the source and target images. This allowed the target images to be viewed as source images without the loss of content. A deep generative correlation alignment network (DGCAN) was proposed in \cite{51} to bridge domain discrepancy between real and synthetic images. This was done by applying the content loss to several different layers.\par
Person re-ID intrinsically requires domain adaption, since, a model trained on a labeled dataset consisting of groups of individuals is required to perform well on unlabelled datasets without supervised fine-tuning. Self-supervised domain adaptation was used in \cite{23} for person re-ID. The source data was pre-processed using GAN and a baseline was trained in a supervised manner using the pre-processed data. The authors then assigned pseudo labels to target data using the trained model and this constructed dataset was trained in a supervised learning manner. Xiao et al. \cite{24} proposed the domain guided dropout algorithm that discards ineffective neurons while re-identifying individuals on several datasets. In \cite{31}, style based mapping from target domain to source domain was learned between different camera views. An effective and generalized re-ID model was proposed in \cite{33} by investigating the intra-domain variations of the target domain.\par
Our work uses the state-of-the-art image-to-image translation model proposed by Gatys et al. in \cite{50} for transferring styles between first-person and fixed camera dataset.

\section{Approach}

\begin{figure}[h]
    \centering
    \includegraphics[scale = 0.33]{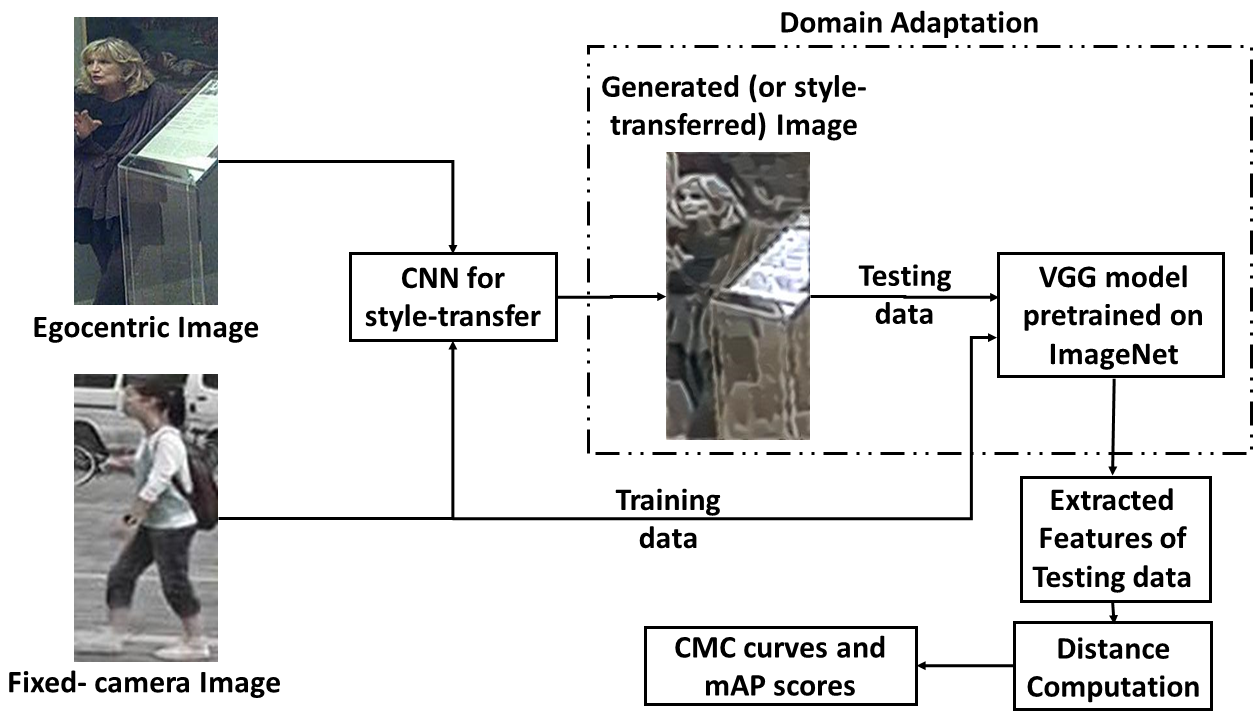}
    \caption{Flow Diagram of our proposed egocentric re-ID model showing various blocks including style transfer and domain adaptation.}
    \label{fig:block}
\end{figure}

The schematic of the proposed egocentric re-ID model is shown in Fig. \ref{fig:block}. The first step in our model is the generation of images which has combined features of fixed camera dataset (Market-1501 \cite{market}) and first-person vision dataset (EGO Re-ID \cite{21}). Our model makes use of a VGG-16 \cite{vgg} based Neural Style Transfer 
(NST) model to produce high-quality images in a manner that helps in bringing the qualities of Market-1501 \cite{market} into EGO Re-ID \cite{21} while preserving the content of EGO Re-ID \cite{21}. This is achieved by formulating image generation problem as an optimization one that minimizes the following losses: \textbf{(1)} Content Loss, \textbf{(2)} Style Loss and \textbf{(3)} Total Loss.

\begin{figure}[H]
    \centering
    \includegraphics[scale = 0.42]{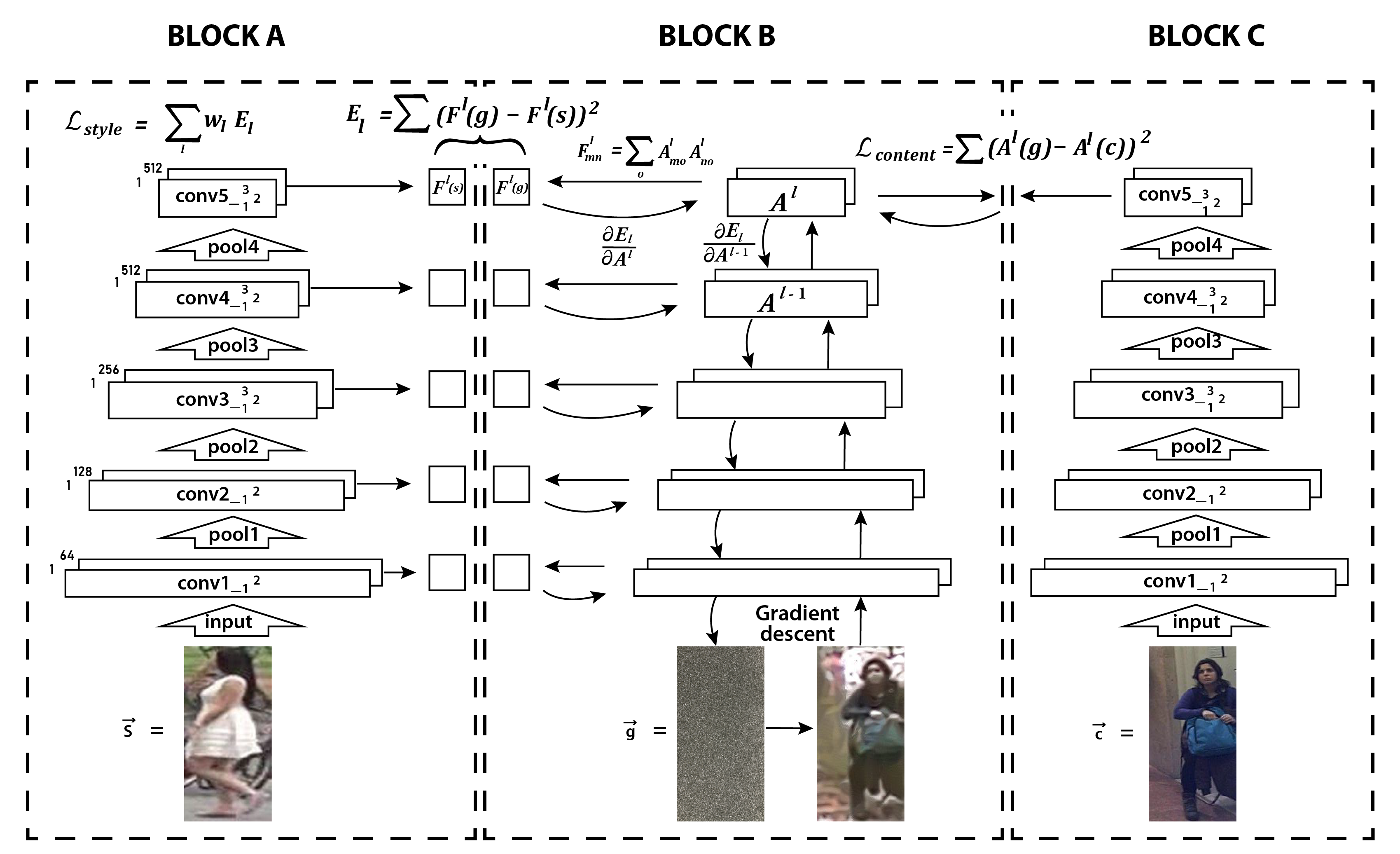}
    \caption{Detailed illustration of Style Transfer Algorithm \cite{50}. Blocks A and B are used for computing style loss between style image and the generated image, Blocks B and C are used for computing content loss between content image and the generated image.}
    \label{fig:sta}
\end{figure}

The content difference between the generated and the EGO Re-ID \cite{21} images are captured through the content loss using a CNN. 
The input image is transformed along the processing hierarchy of the CNN such that the image representations are oblivious to the precise appearance of the image but are aware of the image's actual content. Thus, a network's higher layers do not worry about the exact pixel values. To capture the content information we make use of these image representations which are present at higher layers as shown in Block C of Fig. \ref{fig:sta}\par
The style loss captures the difference between the generated and the Market-1501 \cite{market} images. The feature space responsible for capturing texture information is used for the computation of style loss. By computing the feature correlations between different filter responses, we obtain the style representation of the image. This style representation can thus be used to compute the style loss. Unlike content loss, the style loss is computed across all the layers as shown in Block A of Fig. \ref{fig:sta}\par
After computing the content and style loss, the total loss is calculated as a linear relationship between content and style loss. After the images are generated by minimizing the total loss, a pre-trained VGG model is fine-tuned using images from Market-1501 \cite{market} and image features are calculated for these generated images after passing them through this fine-tuned model employing domain adaptation as shown in the flow diagram of Fig. \ref{fig:block}.
Furthermore, average pooling
operation is adopted for the image synthesis task. According to \cite{50}, replacing the max pooling with an average pooling offer somewhat better results.\par

\subsection{Content Loss}
Block C in Fig. \ref{fig:sta} shows the computation of content loss. 
Let $\vec{g}$ and $\vec{c}$ be the generated and content image respectively. And $A_{m n}^{l}(I)$ be the feature representations of the $m^{th}$ feature map at position \textit{n} in layer \textit{l} of image \textit{I}. Hence, the squared error loss between the content image and the generated image is:
\begin{equation}\label{eq:contentloss}
\mathcal{L}_{\text {content }}=\frac{1}{2} \sum_{m, n}\left(A_{m n}^{l}(g)-A_{m n}^{l}(c)\right)^{2}
\end{equation}
The loss is back-propagated by taking its derivative with the activations present in layer \textit{l} as shown in Block B in Fig. \ref{fig:sta}. And the initial white noise image $\vec{g}$ keeps on changing until we get the same response as the content image $\vec{c}$.\par

\subsection{Style Loss}
Block A in Fig. \ref{fig:sta} shows the computation of style loss. The feature map correlations are given by Gram matrix:
\begin{equation}\label{eq:gram}
F_{m n}^{l}(I)=\sum_{o} A_{m o}^{l}(I) A_{n o}^{l}(I)
\end{equation}
where, $F_{m n}^{l}(I)$ gives the inner product between the feature maps (in vectorised form) \textit{m} and \textit{n} at layer \textit{l} of image \textit{I}.\par

Let $\vec{s}$ and $\vec{g}$ be the original and generated images respectively. And $F_{m n}^{l}(I)$ be their feature representations of the $m^{th}$ filter which is located at position \textit{n} in the layer \textit{l} of image \textit{I}. Hence, the squared error loss between the style image and generated image is:
\begin{equation}\label{eq:styleloss}
E_{l}=\frac{1}{4 N_{l}^{2} M_{l}^{2}} \sum_{m, n}\left(F_{m n}^{l}(g)-F_{m n}^{l}(s)\right)^{2}
\end{equation}
where, the product of $N_l$ and $M_l$ represents the size of the layer \textit{l}.\par
Since, style loss consists of multiple layers, we need to define a weighting factor in order to incorporate the contribution of each layer to the loss. Thus, giving us the total style loss as:
\begin{equation}
\mathcal{L}_{\mathrm{style}}=\sum_{l=0}^{L} w_{l} E_{l}
\end{equation}
The derivatives of $E_l$ with respect to the activations of layer \textit{l} can be computed and backpropagated as shown in Block B in Fig. \ref{fig:sta}.

\subsection{Total Loss}
We intend to jointly minimize the content loss between the white noise image and the content from the first-person image at one layer and the style loss between the white noise image and the style from the fixed camera image on several layers. The loss function to be minimized is:
\begin{equation}\label{eq:totalloss}
\mathcal{L}_{\text {total }}=\alpha \mathcal{L}_{\text {content }}+\beta \mathcal{L}_{\text {style }}
\end{equation}
where $\alpha$ is the weight factor for content reconstruction while $\beta$ is the weight factor for style reconstruction.\par
There is a trade-off between content and style being represented in a generated image. If the generated image has learned better content then its style will be bad and vice-versa. To balance both content and style, weighting factors are included in the final loss expression.

\subsection{Style Transfer based Person Re-ID}
The style transferred images generated by the NST algorithm are used to re-identify individuals. VGG-16 \cite{vgg} is used to extract the image descriptors. The overfit reduction techniques used were weight decay and dropout. To merge the EGO Re-ID dataset \cite{21} (2343 images of 24 persons) with a fixed camera dataset, we used an equal number of images from the fixed camera dataset (Market-1501 \cite{market}). 
The `Training Set 1' consists of 2343 images from Market-1501 dataset \cite{market} which were used as style images with the EGO Re-ID \cite{21} content images. The `Training Set 2' consists of arbitrary 2343 images from the Market-1501 dataset \cite{market}. These images were not used for style transferring the EGO Re-ID dataset \cite{21}. The `Training Set 3' consists of images from both Training Set 1 and Training Set 2. Thus, comprising a total of 4686 images from Market-1501 dataset \cite{market}.\par
These three sets of images were used to finetune a VGG-16 model pre-trained on ImageNet. The three separate trained models were then used to predict the classes for the style transferred images from the EGO Re-ID dataset \cite{21}. The results obtained through various evaluation methods are presented in Table \ref{tab:valueall}.\par
For predicting the classes of the individuals from the first-person dataset, some changes were made to the trained model. The final softmax layer of the models trained on Training Set 1, Training Set 2, and Training Set 3 image sets were removed. So, after passing the images through the trained model, a 4096 length vector was extracted from the fifteenth layer of the network. This vector comprises all the features present in the style transferred images and is referred to as the image descriptor. The distance between these image descriptors was calculated using the Euclidean metric. Also, CMC curves are generated based on the Euclidean distance between these extracted feature vectors. Since there are multiple ground truths for a single query image, mAP scores were also computed.\par

\section{Experiments and Results}
\subsection{Datasets}
The experiment was conducted using two datasets to evaluate our model, (1) Market-1501 \cite{market} as style dataset, and (2) EGO Re-ID \cite{21} as content dataset.\par
The Market-1501 dataset \cite{market} consists of over 32,000 annotated bboxes with over 1500 identities, plus a distractor set of over 500K images
\par
EGO Re-ID dataset \cite{21} is 
recorded in 3 different locations through a frontal camera of an eye-tracking device. 
Each location consists of 8 persons with 100 images of each under different viewpoints. There are a total of 2343 images across 24 identities.

\subsection{Evaluation Methodologies}
Cumulative matching characteristic (CMC) curves are most commonly used for evaluating person re-ID algorithms. 
They are accurate for evaluating re-ID methodology when only one ground truth for each query exists. In this work, a query can have multiple ground truths, so using only CMC curves does not fully reflect the true ranking results as well as the accuracy of the model. To tackle this issue we use both CMC and mAP (mean Average Precision) scores as the evaluation metric.

\subsection{Implementation Details}
TensorFlow was used to build our model. Style loss is calculated across layers 1\textunderscore1, 2\textunderscore1, 3\textunderscore1, 4\textunderscore1 and 5\textunderscore1 of VGG-16 as shown in Block A of Fig. \ref{fig:sta}. For content loss, layer 5\textunderscore2 was used as shown in Block C of Fig. \ref{fig:sta}. 
Each layer contributed equally to the style loss, hence the value of $w_{l}$ for all the layers was set at 0.2. The ratio of $\alpha$ and $\beta$ was set to 1 $\times$ $10^{-3}$ in the final loss function. The L2 regularization was set to a value of 5 $\times$ $10^{-4}$ in VGG to reduce overfitting of data. Also, the dropout rate was set to 0.5. 100 epochs were performed to generate style transferred images using NST. The learning rate was set to 1 $\times$ $10^{-3}$. Each image took 3-4 minutes to generate. During training Adam optimizer was used with its default parameter values, i.e., $\beta_1$ = 0.9 and $\beta_2$ = 0.999. 

\subsection{Results}
\begin{figure}[h]
\begin{subfigure}{.3\textwidth}
  \centering
  \includegraphics{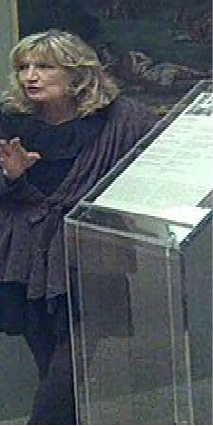}
  \caption{}
  \label{fig:sfig1}
\end{subfigure}
\begin{subfigure}{.3\textwidth}
  \centering
  \includegraphics{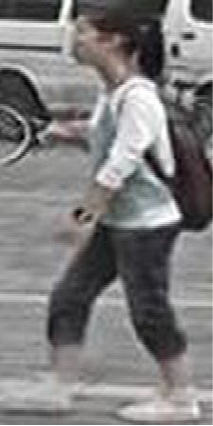}
  \caption{}
  \label{fig:sfig2}
\end{subfigure}
\begin{subfigure}{.3\textwidth}
  \centering
  \includegraphics{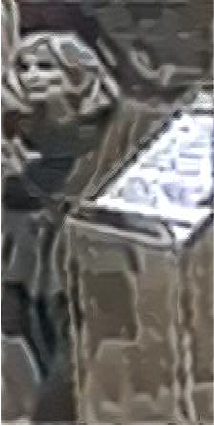}
  \caption{}
  \label{fig:sfig2}
\end{subfigure}
\caption{Success Case of image generated by NST: (a) Content image (EGO Re-ID dataset \cite{21}); (b) Style image (Market-1501 dataset \cite{market}); (c) Style transferred image. The incorporation of style into the content is very well.}
\label{fig:three}
\end{figure}

\begin{figure}[h]
\begin{subfigure}{.3\textwidth}
  \centering
  \includegraphics{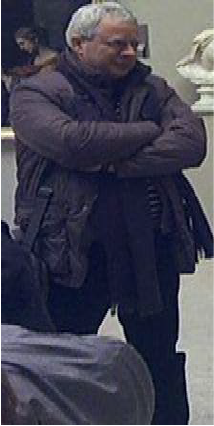}
  \caption{}
  \label{fig:sfig1}
\end{subfigure}
\begin{subfigure}{.3\textwidth}
  \centering
  \includegraphics{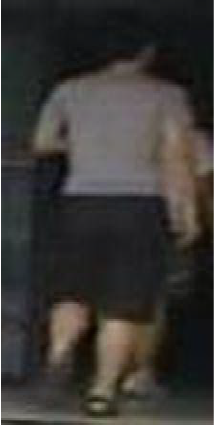}
  \caption{}
  \label{fig:sfig2}
\end{subfigure}
\begin{subfigure}{.3\textwidth}
  \centering
  \includegraphics{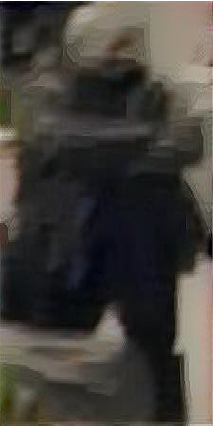}
  \caption{}
  \label{fig:sfig2}
\end{subfigure}
\caption{Failure Case of image generated by NST: (a) Content image (EGO Re-ID dataset \cite{21}); (b) Style image (Market-1501 dataset \cite{market}); (c) Style transferred image. The content and generated images are similar except for the blur due to similarity in content and style images. Major reason for the loss of perceptual quality can be attributed to the fact that the style images are of low quality, hence the quality of all the generated images is degraded to an extent.}
\label{fig:seven}
\end{figure}

Given a content and style image, NST generated flawless images which incorporates both style and content images. A large number of generated images were successful in blending content and style (Fig. \ref{fig:three}) whereas only a few images were unsuccessful in blending the styles (Fig. \ref{fig:seven}). These failure cases can be attributed to the lack of proper transfer of content or style features from the content and style images respectively.

The training data for our work was divided into three sets as mentioned earlier. The details of the nomenclature used are provided below:
\begin{itemize}
    \item Basic Model is the VGG model trained on the ImageNet dataset. No other fine-tuning was done on this model.
    \item Training Set 1, 2, and 3 were used to fine-tune the Basic Model to obtain Model 1, 2, and 3 respectively. 
    \item Style transferred images are the images generated by the NST algorithm.
    \item Normal images are the images from EGO Re-ID dataset \cite{21} without undergoing style transfer.
\end{itemize}
The details mentioned above about different models will be helpful in understanding the contents of Table \ref{tab:valueall}.\par
From Table \ref{tab:valueall}, we can observe that the style transferred images whose features were extracted through Model 1 has the highest mAP score and the highest recognition rate across all ranks. We can also observe that the recognition rate of normal images across all the training models when compared with their style-transferred counterparts are the lowest.

\begin{table}[h]
    \centering
    \caption{Column 1 lists the training sets which were used. Recognition percentage (\%) at r-1, r-5 and r-10 are provided for different training sets. The mean Average Precision (mAP) scores of model trained under different training sets are provided in the last column.}
    \label{tab:valueall}
    \begin{tabular}{|c|c|c|c|c|c|}\hline
    \textbf{Training Model} & \textbf{Style Transferred} &\textbf{Rank 1} & \textbf{Rank 5} & \textbf{Rank 10} & \textbf{mAP} \\\hline\hline
     \multirow{2}{*}{\textbf{Model 1}}   & Yes & 57.66 & 75.07 & 81.63 & 65.88 \\\cline{2-6}
          & No & 19.60 & 43.74 & 59.32 & 31.79 \\\hline
    \multirow{2}{*}{\textbf{Model 2}}   & Yes & 34.47 & 55.12 & 66.23 & 44.32 \\\cline{2-6}
          & No & 26.60 & 45.67 & 56.26 & 36.35 \\\hline
    \multirow{2}{*}{\textbf{Model 3}}   & Yes & 44.53 & 65.70 & 75.33 & 55.22 \\\cline{2-6}
          & No & 18.98 & 44.27 & 59.23 & 31.61 \\\hline
     \multirow{2}{*}{\textbf{Basic Model}}   & Yes & 33.42 & 62.12 & 75.15 & 47.30 \\\cline{2-6}
          & No & 29.22 & 47.94 & 59.32 & 38.85 \\\hline
    \end{tabular}
\end{table}
The high values of recognition rate obtained after computing the distance between the features of style transferred images extracted from Model 1 implies that individuals in first-person domain can be re-identified through domain adaptive techniques. For rank-1, the recognition percentage has jumped from 18.98\% when using Model 3 and non-style transferred images to 57.66\% when Model 1 and style transferred images are utilised. This is a 203.8\% increment in the recognition rate. Even when considering the same model, i.e., Model 1, the recognition rate for rank-1 has jumped from 19.6\% to 57.66\% when non-style transferred images were replaced by style transferred images, which is a 194.2\% increment in recognition percentage. Similar trends can be observed across all the ranks.\par
The main aim of this work was to present a egocentric person re-ID algorithm with good accuracy and a technique to eliminate dataset bias so that the models trained on third-person dataset does not become obsolete while testing on a first-person dataset. From Table \ref{tab:valueall} it is clear that this re-ID algorithm has good accuracy and the increment in recognition percentage when non-style transferred images are replaced with style transferred images is significant which shows that even though the model is trained on a third-person dataset, it's quite accurate when tested on a first-person dataset. The first-person dataset used in this work has been used for a different re-ID algorithm as well \cite{21}. The comparison of the mAP scores of both the models has been provided in Table \ref{tab:compare}.\par

\begin{table}[h]
    \centering
    \caption{Comparison of mean Average Precision (mAP) scores between \cite{21} and our approach}
    \label{tab:compare}
    \begin{tabular}{|c|c|}\hline
    \textbf{Approach}     & \textbf{mAP scores} \\\hline\hline
    Using Torso description & 53.1\\\hline
    Using Face description & 60.3 \\\hline
    Using Full body (with uniform weights) & 67.5\\\hline
    Using Full body (with gaze weights) & 70.3\\\hline\hline
    \textbf{Ours} & \textbf{65.88}\\\hline
    \end{tabular}
\end{table}
From Table \ref{tab:compare} we can infer that our approach has better mAP scores when a few body parts of the individuals are used for re-ID in \cite{21}. But when the full body of the individual is considered for re-ID in \cite{21}, our approach falls behind, but not by a large margin. Even though, our mAP score is lower than the approach in \cite{21} when considering full body of the individual, our approach utilizes style transfer, which can give consistent results irrespective of the dataset it is trained on. The approach in \cite{21} will fall behind in mAP scores when it is tested on a different dataset due to the issue of dataset bias. Our work can thus perform superior irrespective of datasets it will be trained on.\par
\begin{table}[h]
    \centering
    \caption{Recognition percentage (\%) at r-1 and r-5 are provided to compare our approach with  different state-of-the-arts approaches. The mean Average Precision (mAP) scores of our approach and different state-of-the-arts approaches are provided in the last column.}
    \label{tab:compareall}
    \begin{tabular}{|c|c|c|c|}\hline
        \textbf{Approach} & \textbf{Rank-1} & \textbf{Rank-5} & \textbf{mAP scores} \\\hline\hline
    PSE+ECN \cite{55} & 15.17 & 25.79 & 8.58\\\hline
    MGCAM \cite{56} & 18.48 & 29.79 & 14.60 \\\hline
    EgoRe-ID \cite{18} & 53.02 & 63.52 & 44.79\\\hline\hline
    \textbf{Ours} & \textbf{57.66} & \textbf{75.07} & \textbf{65.88}\\\hline
    \end{tabular}
\end{table}
There has been other works on egocentric person re-ID using a different ego re-ID dataset. The comparison of the results of all those work with ours are provided in Table \ref{tab:compareall}. The datasets used in the state-of-the-art approaches mentioned in Table \ref{tab:compareall} are different from our dataset. EgoRe-ID dataset is much bigger than EGO Re-ID dataset \cite{21}. The difference in our results can be attributed to the differences in sizes of the dataset. Overall our approach performed well as can be seen from Table \ref{tab:compare} and \ref{tab:compareall}.

\section{Conclusion}
We have proposed a novel egocentric person re-ID approach by transferring styles of fixed camera datasets into first-person datasets. Fixed camera approaches cannot be used for re-identifying when the dataset changes due to dataset bias. Extensive experimentation by using different sets of training data has shown that our approach of re-identifying individuals through domain adaptation is acceptable. Increment in recognition rate as high as 203.8\% was observed when style-transferred images were used. The best model came out to be Model 1, which was trained on the same images from Market-1501 \cite{market} which were used for style transferring the EGO Re-ID dataset \cite{21}. NST algorithm learns the style features from the Market-1501 dataset \cite{market} in a really good way with some exceptions. The computation time required to generate these images is the bottleneck. Although our model performs well, some of the images generated by NST are not incorporating elements of the fixed camera dataset as discussed in failure cases. This problem can be solved by using GAN based image translation algorithms.

%
%
\bibliography{my}
\bibliographystyle{splncs04}





\end{document}